\def\year{2018}\relax
\documentclass[letterpaper]{article} 
\usepackage{aaai18}  
\usepackage{times}  
\usepackage{helvet}  
\usepackage{courier}  
\usepackage{url}  
\usepackage{graphicx}  
\usepackage{caption}
\usepackage{subcaption}
\usepackage{wrapfig}
\usepackage{tabularx}
\usepackage{placeins}
\usepackage{dblfloatfix}
\usepackage[table]{xcolor}

\usepackage{algorithm}
\usepackage{algpseudocode}
\usepackage{pifont}

\usepackage{amsmath}
\usepackage{amssymb}
\usepackage{amsfonts}
\usepackage{amsthm}
\usepackage{mathtools}
\usepackage{bm}
\usepackage{mathrsfs}
\usepackage{color}
\usepackage{verbatim}

\usepackage[hang,flushmargin]{footmisc}   

\def\x{\mathbf{x}}

\def\a{\mathbf{a}}

\def\p{\mathbf{p}}

\def\z{\mathbf{z}}

\def\R{\mathbb{R}}

\def\balpha{\boldsymbol{\alpha}}

\definecolor{mycolor}{cmyk}{0.0,0.2,0.9,0.0}

\begin{document}
%
\title{ Joint Dictionaries for Zero-Shot Learning}
\author{Soheil Kolouri \thanks{Equal contribution} \\
  HRL Laboratories, LLC\\
  \texttt{\small skolouri@hrl.com} \\
  \And
  Mohammad Rostami~$^*$\\
  University of Pennsylvania\\
  \texttt{\small mrostami@seas.upenn.edu} \\
  \And 
  Yuri Owechko\\
  HRL Laboratories, LLC\\
  \texttt{\small yowechko@hrl.com} \\
  \And
  Kyungnam Kim\\
  HRL Laboratories, LLC\\
  \texttt{\small kkim@hrl.com}}
 \maketitle

\begin{abstract}

A classic approach toward zero-shot learning (ZSL) is to map the input domain to a set of semantically meaningful attributes that could be used later on to classify unseen classes of data (e.g. visual data). In this paper, we propose to learn a visual feature dictionary that has semantically meaningful atoms. Such dictionary is learned via joint dictionary learning for the visual domain and the attribute domain, while enforcing the same sparse coding for both dictionaries. Our novel attribute aware formulation provides an algorithmic solution to the domain shift/hubness problem in ZSL. Upon learning the joint dictionaries, images from unseen classes can be mapped into the attribute space by finding the attribute aware joint sparse representation using solely the visual data.  We demonstrate that our approach provides superior or comparable performance to that of the state of the art on benchmark datasets. 
\end{abstract}

\section{Introduction}

Most classification algorithms require a large pool of manually labeled data to learn the optimal parameters of a classifier. The recent exponential growth of visual data, the growing need for fine-grained multi-label annotations, and consistent emergence of new classes (e.g. new products), however, has rendered manual labeling of data practically infeasible. Transfer learning has been proposed as a remedy to deal with this issue \cite{lampert2014attribute}. The idea is to learn on a limited number of classes and then through knowledge transfer, learn how to classify images from the new classes either using only few labeled data points, i.e. few- and one-shot learning \cite{fei2006one}, or in the extreme case without any labeled data, i.e. zero-shot learning (ZSL) \cite{lampert2014attribute}.  These transfer learning approaches  address the challenge of annotated data unavailability and open the door towards lifelong learning machines. 

To learn target classes with no labeled data, one needs to be able to generalize the relationship between the source data and its labels to the target classes.  To address this challenge in ZSL, an intermediate shared space (i.e. the space of semantic attributes) is exploited, which allows for knowledge transfer from labeled classes to the unlabeled classes. 
The overarching idea in ZSL is that the source and the target classes share common  attributes. The semantic attributes (e.g., can fly, is green) are often provided as accessible side information (e.g. verbal description of a class), which uniquely describe classes of data. To achieve ZSL the relationship between  seen data and its corresponding attributes are first learned in the training phase. In testing stage, this allows for parsing a target image from an unseen class into its semantic attributes to predict  corresponding label.

\begin{figure} 
        \includegraphics[width=\columnwidth]{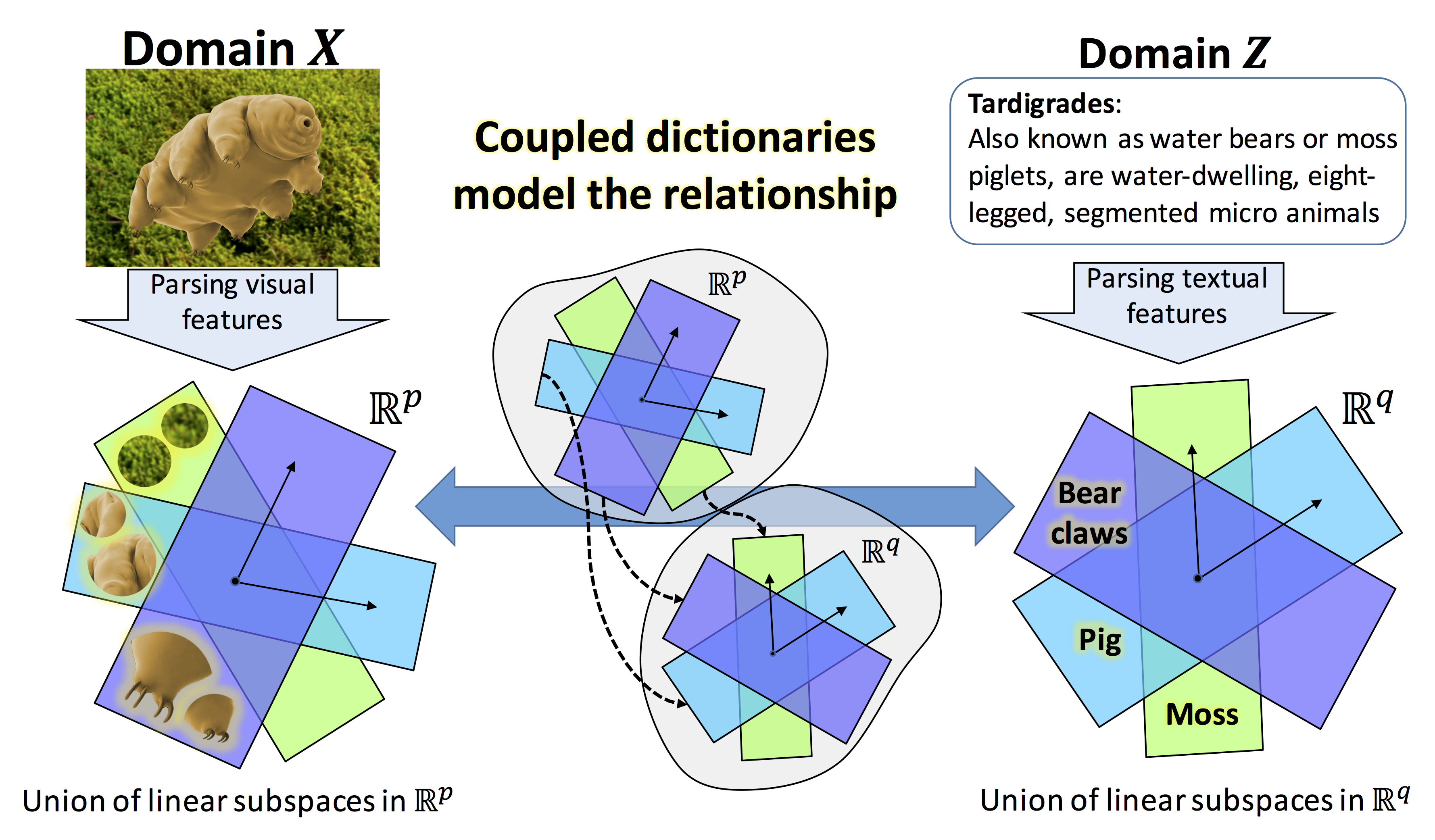}      
   \caption{High-level overview of our  approach. Left \& right: visual and attribute feature extraction and   representation using union of subspaces. Middle:  constraining the dictionary atoms to be coupled. }
\label{fig:tardigrade}
\end{figure}

To clarify the  ZSL core idea and the required steps to perform ZSL, consider the following sentence: `Tardigrades (also known as water bears or moss piglets) are water-dwelling, eight-legged, segmented micro animals'\footnote{Source: Wikipedia}. Given this textual description, one can easily identify the creature in Figure \ref{fig:tardigrade}, Left as a Tardigrade even though she may have never seen one before. Performing this task requires three capabilities: 1) parsing the textual information into semantic features, so we can describe the class {\it Tardigrade} as `bear-like', `piglet-like', `water-dwelling', `eight-legged', `segmented', and `microscopic animal', 2) parsing the image into its visual attributes (See Figure \ref{fig:tardigrade}), and 3) matching the parsed visual features to the parsed textual information which often requires extensive prior knowledge.   Recent textual features extracted from large unlabeled text corpora; including {\it word2vec} \cite{mikolov2013distributed} and {\it glove} \cite{pennington2014glove} enable a learner to efficiently parse textual information. Deep convolutional neural networks (CNNs) \cite{krizhevsky2012imagenet,simonyan2014very,he2016deep,huang2017densely} have revolutionized the field of computer vision and they enable a learner to extract rich visual features from images. An extensive body of work in the field of ZSL is concentrated on modeling the relationship between visual features and semantic attributes
\cite{palatucci2009zero,akata2013label,socher2013zero,norouzi2013zero,lampert2014attribute,zhang2015zero,ding2017lowrank}. 

In this paper, we provide a novel approach to model the relationship between the visual features and the textual information. Our specific contributions are:
\begin{enumerate}
\item New formulation of ZSL via joint dictionary learning
\item Extending the classic joint dictionary learning formulation to an attribute aware formulation that addresses the domain shift/adaptation problem \cite{kodirov2015unsupervised}
\item Demonstrating the benefit of a transductive learning scheme to reduce the hubness phenomenon \cite{dinu2014improving,shigeto2015ridge} 
\end{enumerate}

\section{Related Work}     


ZSL methods often focus on learning the relationship between the visual space and the semantic attribute space. Palatucci et al. \cite{palatucci2009zero} proposed to learn a linear compatibility between the visual space and the semantic attribute space. Lampert et al. \cite{lampert2014attribute} posed the problem as an attribute classification problem and learned individual binary  attribute classifiers in the training stage and used the ensemble of classifiers to map visual features to their  semantic attributes.   
Yu  and Aloimonos \cite{yu2010attribute} approached the problem from a probabilistic point of view and proposed to use generative models to learn prior distributions for image features with respect to each attribute. More recently,  various authors have proposed to embed image features and semantic attributes in a shared metric space (i.e. a latent embedding) \cite{akata2013label,romera2015embarrassingly,zhang2015zero} while  forcing the embedded representations for image features and their corresponding semantic attributes to be similar.  Akata et al. \cite{akata2013label}, for instance,  proposed a model that embeds the image features and the semantic attributes in a common space   (i.e. a latent embedding) where the compatibility between them is measured via a bilinear function.   Similarly, Romera-Paredes and Torr \cite{romera2015embarrassingly} utilized a principled choice of regularizers that enable the authors to derive a simple closed form solution to learn a linear mapping that embeds the image features and the semantic attributes in a low dimensional shared linear subspace.  Others have identified the major problems and challenges in ZSL to be the domain shift problem \cite{kodirov2015unsupervised} and the hubness phenomena \cite{dinu2014improving,shigeto2015ridge}. In short, the domain shift problem raises from the fact that the distribution of features corresponding to the same attribute for seen and unseen images could be very different (e.g. stripes of tigers versus zebras). The hubness problem, on the other hand, states that there will often be attributes that are similar (have small distance) to vastly different visual features in the embedding space.  Various transductive approaches are presented to overcome the hubness problem \cite{fu2015transductive,yu2017transductive}.

The use of sparse dictionaries to model the space of visual features and semantic attributes as union of linear subspaces has  been shown to be an effective modeling scheme in recent ZSL papers \cite{yu2017transductive,isele2016using,kodirov2015unsupervised,zhang2015zero}.  Zhang et al. \cite{zhang2015zero} showed that modeling the test image features as sparse linear combination of train image features is beneficial and formulated a ZSL method based on this principal. Using similar ideas, Isele et. al. \cite{isele2016using} used joint dictionary learning to learn a dynamical control system using high level task descriptors in an online lifelong zero-shot reinforcement learning setting.  Our JD-ZSL build on similar ideas as in \cite{yu2017transductive,isele2016using,kodirov2015unsupervised} and introduce a novel ZSL method based on learning joint sparse dictionaries for the image features and the semantic attributes. At its core, JD-ZSL is equipped with a novel entropy minimization regularizer \cite{grandvalet2004semi}, which facilitates the solution to the ZSL problem by reducing the domain shift effect. We further show that a transductive approach applied to our attribute aware JD-ZSL formulation provide state-of-the-art or close to state-of-the-art performance on various benchmark datasets. Finally it should be noted that the idea of using joint dictionaries to map data from a given metric space to a second related  space was pioneered by Yang et al.  \cite{yang2010image} in super-resolution applications.  

Figure \ref{fig:tardigrade} captures the gist of our idea. Visual features are extracted via CNNs, left sub-figure, and the semantic attributes are provided via textual feature extractors like word2vec or via human annotations, right sub-figure.  Both the visual features and the semantic attributes are assumed to be representable sparsely in a shared union of linear subspaces, left and right sub-figures. The idea here is then to enforce the sparse representation vectors for both domains be equal and thus  effectively couple the learned dictionaries for the the visual and the attribute spaces.  
 The intuition from  a co-view  perspective \cite{yu2014discriminative} is that both the visual and the attribute features provide information about the same class, and so each can augment the learning of the other. Each underlying class is common to both views, and one can find task embeddings that are consistent for both the visual features   and their corresponding attributes. Having learned the coupled dictionaries,   zero-shot classification can be performed by mapping images of unseen classes into the attribute space, where classification can be simply done via nearest neighbor or via a more elaborate scheme like label propagation. Given the coupled nature of the learned dictionaries, an image could be mapped to its semantic attributes by first finding the sparse representation with respect to the visual dictionary, and next  the semantic attribute dictionary can be used to recover the attribute  vector from the joint sparse representation which could then be used for classification. 
 
\section{Problem Statement and Technical Rational}

Consider a visual feature metric space $\mathcal{X}$ of dimension  $p$, an  attribute metric space $\mathcal{Z}$ with dimension $q$, and a class label set $\mathcal{Y}$ with dimension $K$ which ranges over a finite alphabet of size $K$ (images can potentially have multiple memberships to the classes). As an example $\mathcal{X}=\mathbb{R}^p$ for the visual features extracted from a deep CNN and $\mathcal{Z}=\{0,1\}^q$ when a binary code of length $q$ is used to identify the presence/absence of various characteristics in an object \cite{lampert2014attribute}. We are given a labeled dataset $\mathcal{D}=\{((\bf{x}_i;\bf{z}_i),\bf{y}_i)\}_{i=1}^N$ of features of seen images and their corresponding semantic attributes, where $\forall i:\bf{x}_i\in \mathcal{X}, \bf{z}_i\in \mathcal{Z}$, and $\bf{y}_{i}\in \mathcal{Y}$.  We are also given the unlabeled attributes of unseen classes  $\mathcal{D}'=\{\bf{z}'_j\}_{j=1}^M$ (i.e. we have access to textual information for a wide variety of objects but do not have access to the corresponding visual information).  In ZSL the set of seen and unseen classes are disjoint and it is  assumed that the semantic attributes are class specific. The goal is then to use $\mathcal{D}$ and $\mathcal{D}'$ to learn the relationship between $\mathcal{X}$ and $\mathcal{Z}$ so when an unseen image (image from an unseen class) is fed to the system, its corresponding attributes and consequently its label could be predicted. Finally, we assume that $\psi:\mathcal{Z}\rightarrow \mathcal{Y}$ is the mapping between the attribute space and the label space and $\psi$ is a known linear mapping, $\bf{y}=\psi(\bf{z})=V\bf{z}$. 

To further clarify the problem, consider an instance of ZSL in which features extracted from images of horses and tigers are included in seen visual features $X=[\bf{x}_1,...,\bf{x}_N]$, where $\bf{x}_i\in\mathcal{X}$, but $X$ does not contain features from images containing zebras. On the other hand, the semantic attributes contain information of all seen $Z=[\bf{z}_1,...,\bf{z}_N]$ for $\bf{z}_i\in\mathcal{Z}$ and unseen $Z'=[\bf{z}'_1,...,\bf{z}'_M]$ for $\bf{z}'_j\in\mathcal{Z}$ classes including the zebras. Intuitively, by learning the relationship  between the image features and the attributes ``has hooves", ``has mane'', and ``has stripes" from the seen images, we must be able to assign an image of a zebra to its corresponding attribute, while we have never seen a zebra before.   More formally, we want to learn the  mapping  $\phi:\mathcal{X}\rightarrow\mathcal{Z} $ which relates the visual space and the attribute space. Having learned this mapping, for an unseen image one can recover the corresponding attribute vector using the image features and then classify the image using the mapping $\bf{y}=(\psi\circ\phi)(\bf{x})$, where `$\circ$' represents function composition. 

\subsection{Technical Rational}

 For the rest of our discussion we assume that $\mathcal{X}=\R^p$, $\mathcal{Z}=\R^q$, and $\mathcal{Y}=\R^K$. The simplest ZSL approach is to assume that the mapping $\phi:\mathbb{R}^p\rightarrow \mathbb{R}^q$ is linear, $\phi(\x)=W^T\x$ where $W\in \R^{p\times q}$, and then minimize the regression error $\frac{1}{N}\sum_i \|W^T\x_i-\z_i\|^2_2$ to learn $W$. Despite existence of  a closed form solution for $W$, the solution contains the inverse of the covariance matrix of $X$, $(\frac{1}{N}\sum_i (x_ix_i^T))^{-1}$, which requires a large number of data points for accurate estimation. To overcome this problem, various regularizations are considered for $W$. Decomposition of $W$ as $W=P\Lambda Q$, where $P\in\R^{p\times l}$, $\Lambda\in\R^{l\times l}$, $Q\in\R^{l\times q}$, and $l<min(p,q)$ can also be helpful. Intuitively, $P$ is a right linear operator that projects $\x$'s into a shared low dimensional subspace, $Q$ is a left linear operator that projects  $\z$ into the same shared subspace, and $\Lambda$ provides a bi-linear similarity measure in the shared subspace. The  regression problem then can be transformed into maximizing $\frac{1}{N}\sum_i \x_i^TP\Lambda Q\z_i$, which is a weighted correlation between the embedded $\x$'s and $\z$'s. This is the essence of many ZSL techniques including Akata et al. \cite{akata2013label} and Romera-Paredes et al.\cite{romera2015embarrassingly}.  This technique  can be   extended to nonlinear mappings using  kernel methods. However, the choice of kernels remains  a challenge.

 On the other side of the spectrum, the mapping $\phi:\R^p\rightarrow \R^q$ can be chosen to be highly nonlinear, as in deep neural networks (DNN). Let a DNN be denoted as $\phi(.;\theta)$, where $\theta$ represents the parameters of the network (i.e. synaptic weights and biases). ZSL can then be addressed by minimizing $\frac{1}{N}\sum_i \|\phi(\x_i;\theta)-\z_i\|^2_2$ with respect to $\theta$. Alternatively, one can nonlinearly embed $\x$'s and $\z$'s in a shared metric space via deep nets, $f(\x;\theta_x):\R^p\rightarrow \R^l$ and $g(\z;\theta_z):\R^q\rightarrow \R^l$, and maximize their similarity measure in the embedded space, $\frac{1}{N}\sum_i f(\x_i;\theta_x)^T g(\z_i;\theta_z)$, as in \cite{lei2015predicting}.  Nonlinear methods  are computationally expensive, require a large training dataset, and can easily overfit to the training data. On the other hand, linear ZSL algorithms are efficient, easy to train, and generalizable but they are often outperformed by nonlinear methods.  As a compromise, we model nonlinearities in data distributions as union of linear subspaces with coupled dictionaries. By jointly learning the visual and attribute dictionaries, we effectively model the relationship between the metric spaces. This allows a nonlinear scheme with a computational complexity comparable to linear techniques.  

\section{Zero Shot Learning using Joint Dictionaries}
\label{sec:jointDL}

Joint dictionary learning has been proposed to couple related features from two   metric spaces \cite{yang2010image,shekhar2014joint}.  Yang et al. \cite{yang2010image} proposed the approach to tackle the problem of image super-resolution and Shekhar et al. \cite{shekhar2014joint} used joint dictionary learning for multimodal biometrics recognition. Following a similar framework, the gist of our approach is to learn the  mapping  $\phi:\mathbb{R}^p\rightarrow \mathbb{R}^q$ through two dictionaries, $D_x\in\mathbb{R}^{p\times r}$ and $D_z\in \mathbb{R}^{q\times r}$  that model $X$ and $[Z,Z']$, respectively, where $r>max(p,q)$. The goal is to find a shared sparse representation (i.e. sparse code) $\a_i$ for $\x_i$ and $\z_i$, such that $\x_i=D_x\a_i$ and $\z_i=D_z\a_i$. Below we describe the training and testing phases of our proposed method.

\subsection{Training phase}

The standard dictionary learning is based on minimizing the empirical average estimation error $\frac{1}{N}\|X-D_xA\|^2_F$ on a given training set $X$, where $\ell_1$ regularization on $A$ enforces sparsity:
\begin{eqnarray}
\begin{split}
 D^*_x,A^* =  & \operatorname*{argmin}_{D_x,A} \frac{1}{N}\|X-D_xA\|^2_F+\lambda\|A\|_1  \\
 &\text{s.t.}~ \|D^{[i]}_x\|^2_2 \leq 1.
  \end{split}
\label{eq:mainDx}
\end{eqnarray}  
Here $\lambda$ is the regularization parameter, which controls the sparsity of $A$, and $D_x^{[i]}$ is the i'th column of $D_x$. Alternatively, following the Lagrange multiplier technique, the Frobenius norm of $D_x$ could be used as a regularizer in place of the costraint.

 In our joint dictionary learning framework, we aim to learn $D_x$ and $D_z$ such that they share the sparse coefficients $A$ to represent the seen visual features $X$ and their corresponding attributes $Z$, respectively. An important twist here is that the attribute dictionary, $D_z$, is also required to sparsify the semantic attributes of other (unseen) classes, $Z'$. To obtain such coupled dictionaries we propose the following optimization,
\begin{eqnarray}
\begin{split}
 \underset{D_x,A,D_z,B}{\operatorname*{argmin}} & \{ \frac{1}{Np}(\|X-D_xA\|^2_F+ \frac{p\lambda}{r}\|A\|_1) + \\& \frac{1}{Nq}\|Z-D_zA\|^2_F+   
  \frac{1}{Mq}(\|Z'-D_zB\|^2_F+\\& \frac{q\lambda}{r}\|B\|_1) \} 
  \hspace{.2in} \text{s.t.:}  \|D^{[i]}_x\|^2_2\leq 1,~\|D^{[i]}_z\|^2_2\leq 1
\end{split}
\label{eq:maineq}
\end{eqnarray}
The above formulation combines the dictionary learning problem for $X$ and $Z$ by coupling them via $A$, and also enforces $D_z$ to be a sparsifying dictionary (i.e. a good model) for $Z'$. The optimization in Eq \eqref{eq:maineq}, while convex in each individual term, is highly nonconvex in all variables. Following the approach proposed in  \cite{yang2012coupled} we use an Expectation Maximization (EM) like alternation to update dictionaries $D_x$ and $D_z$. To do so, we rewrite the optimization problem into the following two steps:
\begin{enumerate}
\item For a fixed $D_x$ update $D_z$ via the following optimization:
\begin{eqnarray}
\begin{split}
&\underset{D_z,B}{\operatorname*{min}} \frac{1}{Mq}(\|Z'-D_zB\|^2_F+\frac{q\lambda}{r}\|B\|_1)+ \\&\hspace{.3in}\frac{1}{Nq}\|Z-D_zA^*\|_F^2\\& \text{s.t.}~~ A^*=\operatorname*{argmin}_{A} \frac{1}{p}\|X-D_xA\|^2_F+ \frac{\lambda}{r}\|A\|_1,\\& \|D_z^{[i]}\|_2^2\leq 1
 \end{split}
\label{eq:maineq1}
\end{eqnarray}
$A$ is found using a Lasso optimization problem, and FISTA \cite{beck2009fast} is used to update $D_z$ and $B$. 
\item For a fixed $D_z$ update $D_x$ via:
\begin{eqnarray}
\begin{split}
  &\underset{D_x}{\operatorname*{min}} ~~~ \|X-D_xA^*\|^2_F \\
 & \text{s.t.}~~ A^*=\operatorname*{argmin}_{A} \frac{1}{q}\|Z-D_zA\|_F^2+\frac{\lambda}{r} \|A\|_1,\\& \hspace{.2in}\|D_x^{[i]}\|_2^2\leq 1,
 \end{split}
\label{eq:maineq2}
\end{eqnarray}
which involves a Lasso optimization together with a simple regression with a close form solution. 
\end{enumerate}

\subsection{Zero-Shot Prediction of Unseen Attributes}

In the testing phase we are only given the extracted features from unseen images, $X'=[\x'_1,...,\x'_l]\in\R^{p\times l}$ and the goal is to predict their corresponding semantic attributes. Here we introduce a progression of methods, which clarifies the logic behind our method, and enables us to efficiently predict the semantic attributes of the unseen images based on the learned dictionaries in the training phase. 

\subsubsection{Attribute Agnostic Prediction}

The attribute agnostic (AAg) formulation, is the naive way of predicting semantic attributes from an unseen image $\x'_i$. In the AAg formulation, we first find the sparse representation $\balpha_i$ of the unseen image $\x'_i$ with respect to the learned dictionary $D_x$ by solving the following Lasso problem, 
\begin{equation}
\balpha_i=\operatorname{argmin}_\a \frac{1}{p}\|\x_i-D_x\a\|_2^2 +\frac{\lambda}{r}\|\a\|_1.
\label{eq:attrAgn}
\end{equation}
Here, one can simply use $\balpha_i$ and compare it to the sparse codes of the unseen attributes, $\mathbf{b}_j$. In our experiments, however, we found that this approach is not suitable in our JD-ZSL setting as the dictionaries could have redundant atoms that cause two similar image features or attributes to have different sparse codes. Instead, we do the comparison in the attribute space and predict the corresponding attribute via $\hat{z}_i=D_z\balpha_i$. In the attribute-agnostic formulation,  the sparse coefficients are calculated without any information from the attribute space. Not using the information from the attribute space would lead to the domain shift problem, in the sense that there is no guarantee that $\balpha_i$ would reconstruct a meaningful attribute in $\mathcal{Z}$. In other words, $\hat{z}_i=D_z\balpha_i$ could be far from the unseen attributes, $\z'_m$, and therefore could not be assigned to any known attribute with high confidence. To alleviate this problem we progress to  an extended solution, which we denote as the Attribute Aware (AAw) prediction. 

\subsubsection{Attribute Aware Prediction}

In the attribute-aware (AAw) formulation we would like to find the sparse representation $\balpha_i$ to not only approximate the input visual feature, $\x'_i\approx D_x\balpha_i$, but also provide an attribute prediction, $\hat{z}_i=D_z\balpha_i$, that is well resolved in the attribute space and does not suffer from the domain shift problem. Note that, ideally $\hat{\z}_i=\z'_m$ for some $m\in\{1,...,M\}$. To achieve this we define the soft assignment of $\hat{\z}_i$ to $\z'_m$, denoted by $p_m$, using the Student's t-distribution as a kernel to measure similarity between $\hat{\z}_i=D_z\balpha_i$ and $\z'_m$,
\begin{equation}
p_m(\balpha_i)=\frac{(1+\frac{\|D_z\balpha_i-\z'_m\|^2_2}{\rho})^{-\frac{\rho+1}{2}}}{\sum_k (1+\frac{\|D_z\balpha_i-\z'_k\|^2_2}{\rho})^{-\frac{\rho+1}{2}}}
\label{eq:softass} 
\end{equation}
where $\rho$ is the kernel parameter. The choice of t-distribution is due to its long tail and low sensitivity to the choice of kernel parameter, $\rho$.  Ideally, $p_m(\balpha_i)=1$ for some $m\in\{1,...,M\}$ and $p_j(\balpha_i)=0$ for $j\neq m$. The ideal soft-assignment $\p=[p_1,p_2,...,p_M]$ then would be one-sparse and therefore would have minimum entropy. This motivates our attribute-aware formulation, which regularizes the AAg formulation in Equation \ref{eq:attrAgn} with the entropy of $\p$.
\begin{equation}
\begin{split}
\balpha_i=\operatorname{argmin}_\a&\underbrace{\frac{1}{p}\|\x'_i-D_x\a\|_2^2 -\gamma \sum_m p_m(\a)log(p_m(\a))}_{g(\a)} \\&+\frac{\lambda}{r}\|\a\|_1\end{split}
\label{eq:attrAware}
\end{equation}
where $\gamma$ is the regularization parameter for entropy of the soft-assignment probability vector $\p$. Such entropy minimization scheme has been successfully used in several work \cite{grandvalet2004semi,huang2016sparse} whether as a sparsifying regularization or to boost the confidence of classifiers. We note that the entropy regularization enforces the prediction to be close to one of the unseen attributes, but it can potentially backfire in that a low-entropy solution (aligned to a prototype) doesn't necessarily have to be the correct solution. In our experiments, we consistently observed higher performance for the AAw formulation. 

The entropy regularization turns the optimization in Eq. \eqref{eq:attrAware} into a nonconvex problem. In \cite{huang2016sparse}, the authors use a generalized gradient descent approach similar to FISTA to optimize this non-convex problem. We use a similar scheme to optimize the objective function in Eq. \eqref{eq:attrAware}. In short, we relax $g(\a)$  using its quadratic approximation   around the previous estimation of $\a$, $\a_{k-1}$, and update $\a$ as the solution of the following problem 
\begin{eqnarray}
\begin{split}
\bf{a}_k = \operatorname{argmin}_{\bf{a}}&
\frac{1}{2t}\|\bf{a}-(\bf{a}_{k-1}-t\nabla g(\bf{a}_{k-1}))\|_2^2+\\& \frac{\lambda}{r}\|a\|_1
\end{split}
\label{eq:au1}
\end{eqnarray}
Equation \eqref{eq:au1} is a LASSO problem and can be solved efficiently using FISTA. It only remains to compute  $\nabla g$: 
{\small\begin{eqnarray}
\begin{split}
\nabla g(\a) = &  \frac{1}{p}D_x^T(D_x\a-\x')-\\& \frac{\gamma}{(\sum_k l_k(\a))^2} \sum_m\{(1+log(p_m(\a)))\times \\&(\frac{\partial l_m(\a)}{\partial \a}\sum_k l_k(\a)-l_m(\a)\sum_k \frac{\partial l_k(\a)}{\partial \a}) \}
\end{split}
\label{eq:au2}
\end{eqnarray} }
where:
{\small
\begin{equation*}
\begin{split}
&l_m(\bf{a})=(1+\frac{\| D_z{\bf a}-\bf{z}'_m\|_2^2}{\rho})^{-\frac{\rho+1}{2}},\\&\frac{\partial l_m({\bf a})}{\partial{ \bf a}}=-\frac{\rho+1}{\rho}(D_z^T(D_z{\bf a}-\bf{z}))   (1+\frac{\|D_z{\bf a}-\bf{z}'_m\|_2^2}{\rho})^{-\frac{\rho+3}{2}}.
\end{split}
\label{eq:au4}
\end{equation*}}
Due to the non-convex nature of the objective function, a good initialization is needed to achieve a sensible solution. Therefore we initialize $\balpha$ from the solution of the AAg formulation. Finally the corresponding attributes are estimated by $\hat{z}_i=D_z\balpha_i$, for $i=1,...,l$.

\subsection{From Predicted Attributes to Labels}

\begin{figure*}[t]
\centering
\includegraphics[width=\linewidth]{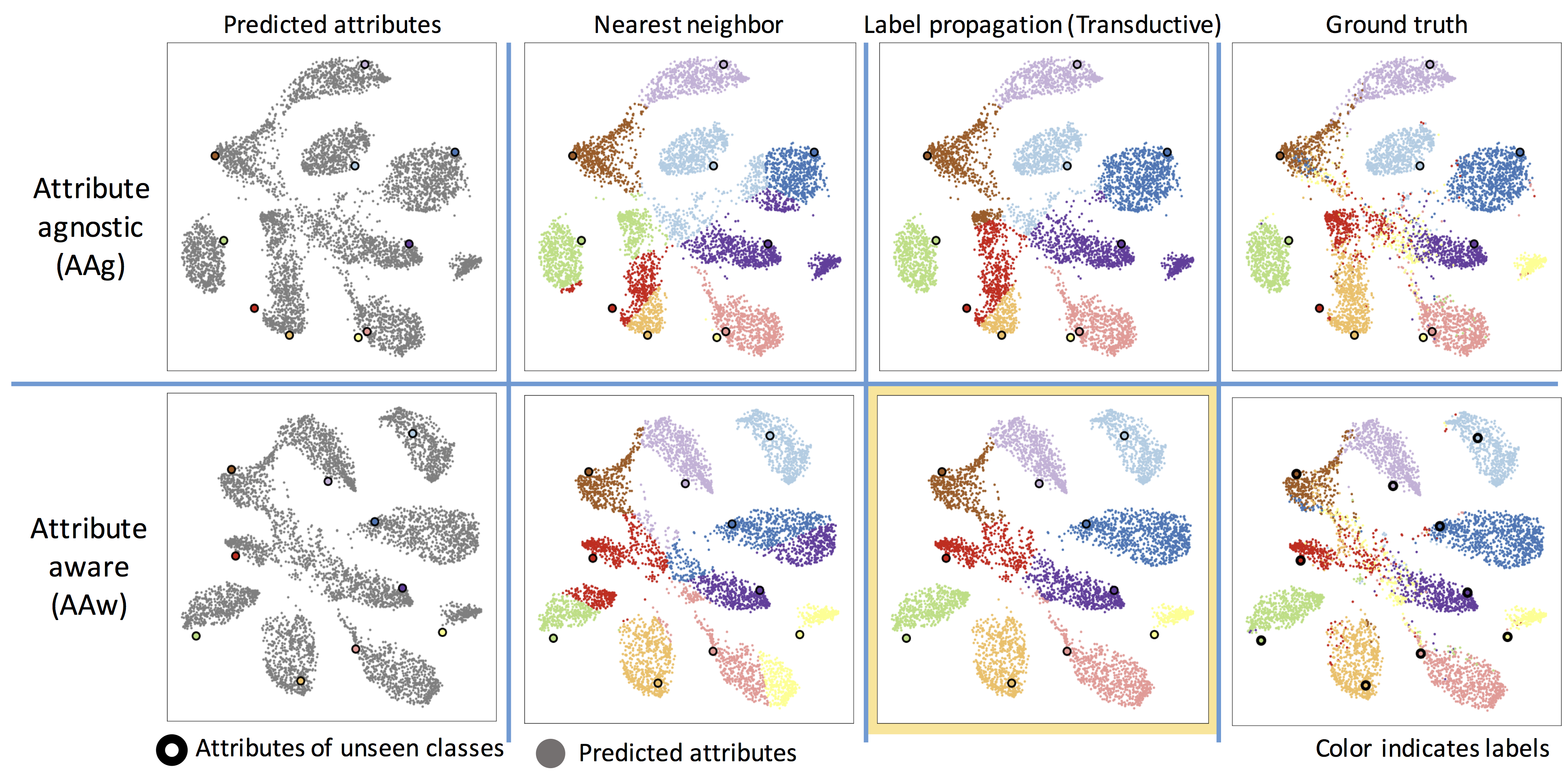}
\caption{Attributes predicted from the input visual features for the unseen classes of images for AWA dataset using our attribute agnostic and attribute aware formulations respectively in top and bottom rows. The nearest neighbor and label propagation assignment of the labels together with the ground truth labels are visualized. It can be seen that the attribute aware formulation together with the label propagation scheme overcomes the hubness and domain shift problems. Best seen in color. }
\label{fig:tsne}
\end{figure*}

In order to predict the image labels, one needs to assign the predicted attributes, $\hat{Z}=[\hat{\z}_1,...,\hat{\z}_l]$, to the $M$ attributes of the unseen classes $Z'$ (i.e. prototypes). This task can be performed in two ways, namely the inductive approach and the transductive approach. In the inductive scheme the inference could be performed using a nearest neighbor (NN) approach in which label of each individual $\hat{\z}_i$ is assigned to be the label of its nearest neighbor $\z'_m$. In such approach the structure of $\hat{\z}_i$'s is not taken into account and the hubness problem could easily degrade the performance of the ZSL algorithm. Looking at the t-SNE embedding  visualization \cite{maaten2008visualizing} of $\hat{\z}_i$'s and $\z'_m$'s in Figure \ref{fig:tsne}, details are explained later, it can be seen that NN does not provide an optimal label assignment. 

In the transductive setting, on the other hand, the attributes for all test images (i.e. unseen) are first predicted to form $\hat{Z}=[\hat{\z}_1,...,\hat{\z}_l]$. Next, a graph is formed on $[Z',\hat{Z}]$ where the labels for $Z'$ are known and the task is to infer the labels of $\hat{Z}$. This problem can be formulated as a graph-based semi-supervised label propagation \cite{belkin2004regularization,zhou2003learning}. We follow the work of Zhou et al. \cite{zhou2003learning} and spread the labels of $Z'$ to $\hat{Z}$. More precisely,  we first reduce the dimension of $[Z',\hat{Z}]$ via t-SNE \cite{maaten2008visualizing} and then form a graph in the lower dimension and perform label propagation on this graph. Figure \ref{fig:tsne} reconfirms that label propagation in a transductive setting could significantly improve the performance of ZSL and resolve the hubness and domain shift issues as also demonstrated in \cite{fu2015transductive,yu2017transductive}.

\section{Theoretical Discussion}
\label{sec:analysis}
The core step for ZSL in our scheme is to compute the joint sparse representation for an unseen image. Note that in the testing phase, the sparse representation $\bf{a}$ is estimated using \eqref{eq:attrAgn}, while the dictionaries are learned for optimal sparse representations as in \eqref{eq:maineq}.   More specifically, we need to demonstrate that  the following two problems lead to close approximations:
\begin{equation}
\begin{split}
\balpha^*&=\operatorname{argmin}_a \|\x-D_x\a\|_2^2 +\|\z-D_z\a\|_2^2+\lambda\|a\|_1\\&= \operatorname{argmin}_a \|\begin{bmatrix}
    \x           \\
    \z           
\end{bmatrix}-\begin{bmatrix}
    D_x           \\
    D_z         
\end{bmatrix}\bf{a}\|_2^2  +\lambda\|\bf{a}\|_1\\
&\balpha^+=\operatorname{argmin}_a \|\x-D_x\a\|_2^2 +\lambda\|\a\|_1,   
\end{split}
\label{eq:dic}
\end{equation}  
in order to conclude that we can solve  for $\balpha^+$ in ZSL regime (i.e. prediction attributes for unseen images) to estimate $\balpha^*$ with good accuracy. Note that the major challenge in the testing phase is that we are using the dictionary $D_x\in\mathbb{R}^{p\times r}$ to find the shared sparse parameters, $\balpha$, instead of $\tilde{D}=[D_x,D_z]^T\in\mathbb{R}^{(p+q)\times r}$. To study the effect of this change, we first point out that  Eq.~\ref{eq:mainDx} can  be interpreted as result of  a maximum a posteriori (MAP) inference from a Bayesian perspective. This means that from a probabilistic perspective, $\balpha$'s are drawn from a Laplacian distribution  and the dictionary $D$ is  a  Gaussian matrix with elements drawn i.i.d: $d_{ij} \sim  \mathcal{N}(\bm{0}, \sigma^2)$. This means that given a drawn dataset, we learn MAP estimate of a Gaussian matrix. In order to analyze the effect, we rely on the following theorem about LASSO with Gausian matrices \cite{negahban2009unified}:

\textbf{Theorem 1 \cite{negahban2009unified}}:
Let $\balpha_s$ be the unique sparse solution of the linear system $\x=D\a$ with $\|\a\|_0=k$ and $D \in \R^{p\times n}$. If $\balpha^\ddagger$ is the LASSO solution
for the system from noisy observations, then with high probability: \mbox{$\|\balpha_s-\balpha^\ddagger\|_2 \leq c'\sqrt{k\frac{\log r}{p}}$}  \enspace, 
where  $c'\in\mathbb{R}^+$ is a constant which depends on the loss function which measures the data fidelity, here the Euclidean distance.

\textbf{Lemma 1}: Attribute prediction error in ZSL setting  is  upper-bounded proportional to $(\frac{1}{\sqrt{p}}+\frac{1}{\sqrt{q+p}})$.

 Proof:  note that if $\balpha^*$ is a solution of $[\bf{x}^T,\bf{z}^T]^T=\tilde{D}\bf{a}$, trivially it is also a solution for $ \x =D_x\a$ as well. Now using Theorem 1:
\begin{equation}
\begin{split}
&\|\z^*-\z^+\|\le \|D_x(\alpha^*-\alpha^+)\|\\
& \|D_x(\alpha^*-\alpha^+)\| \le c'\|D_z\|_2 \sqrt{k\log r}(\frac{1}{\sqrt{p}}+\frac{1}{\sqrt{q+p}})
\end{split}
\label{eq:error}
\end{equation}    
 Note we have used the triangular inequality first and then the theorem in the above deduction and  $\|\cdot\|_2$ denotes  spectral norm for a matrix. This result accords with intuition. First, it advises sparseness of $\z$, i.e. smaller $k$, decreases the error which means that a good sparsifying dictionary would lead to less ZSL error. Second, the error is proportional to inverse of both $\sqrt{p}$ and $\sqrt{p+q}$, meaning that rich visual and attribute descriptions lead to minimal ZSL error. This suggests that for our approach to work, existence of a good sparsifying dictionary as well as rich  visual and attribute data is essential. Finally, although increasing the number of dictionary columns $r$ intuitively can improve sparsity, i.e. decrease $k$, this result shows that it can potentially increase the ZSL error, and should be tuned for an optimal performance.

\begin{table}[t!]
{\small
\begin{tabular}{lc|ccc }   
\multicolumn{2}{c}{Method}    & SUN & CUB & AwA    \\
\hline
\multicolumn{2}{c|}{\cite{romera2015embarrassingly}$^\ddagger$} &  82.10 & - &  75.32 	\\
\multicolumn{2}{c|}{\cite{zhang2015zero}$^\dagger$}	& 82.5 & 30.41 & 76.33 	\\
\multicolumn{2}{c|}{\cite{zhang2016zero}$^\dagger$}      &      82.83   &   42.11  & 80.46\\
\multicolumn{2}{c|}{\cite{bucher2016improving}$^\dagger$}&84.41& 43.29 &  77.32\\
\multicolumn{2}{c|}{\cite{xu2017matrix}$^\dagger$}& 83.5 &  53.6 & 84.5   \\
\multicolumn{2}{c|}{\cite{li2017zero} $^\dagger$}		& - & 61.79 & 87.22   \\
\multicolumn{2}{c|}{\cite{yezero}$^\dagger$ }		&  85.40 &  57.14 & 85.66   \\
\multicolumn{2}{c|}{\cite{ding2017lowrank}$^\dagger$ }		&  86.0 &  45.2 & 82.8   \\
\multicolumn{2}{c|}{\cite{wang2017zero}$^\dagger$} &- & 42.7&79.8  \\
\multicolumn{2}{c|}{\cite{kodirov2017semantic}$^\dagger$} &91.0 & 61.4 &84.7\\
\hline
Ours & AAg \eqref{eq:attrAgn}              &  82.05 & 35.81 &  77.73  \\
Ours & AAw \eqref{eq:softass}             & 83.22 & 38.36 &   83.33  \\
\rowcolor{mycolor!30}
Ours & Transductive AAw (TAAw)            & 85.90 & 47.12 &   88.23  \\
\hline 
Ours &TAAw hit@3 					             &  94.52 & 58.19 &  91.73  \\
Ours &TAAw hit@5					             &  98.15 & 69.67 &  97.13  \\
\end{tabular}}
\caption{ Zero-shot classification results for three benchmark datasets. All methods use VGG19 features trained on the ImageNet dataset and the original continuous (or binned) attributes provided by the datasets. Here, $\dagger$ indicates that the results are extracted directly from the corresponding paper, $\ddagger$ indicates that the results are reimplemented with VGG19 features, and $-$ indicates that the results are not reported. }
\label{tab:table1}
\end{table}

\section{Experiments}
\label{sec:results}
We carried out experiments on three benchmark ZSL datasets and empirically evaluated the resulting performance against  nascent ZSL  algorithms.

{\bf Datasets:} We conducted our experiments on three benchmark datasets namely: the Animals with Attributes (AwA1) \cite{lampertattribute}, the SUN attribute \cite{patterson2012sun}, and the Caltech-UCSD-Birds 200-2011 (CUB)  bird \cite{wah2011caltech} datasets. The AwA1 dataset is a coarse-grained dataset containing 30475 images of 50 types of animals with 85 corresponding attributes for these classes. Semantic attributes for this dataset are obtained via human annotations. The images for the AWA1 dataset are not publicly available; therefore we use the publicly available features of dimension $4096$ extracted from a VGG19 convolutional neural network, which was pretrained on the ImageNet dataset. Following the conventional usage of this dataset, 40 classes are used as source classes to learn the model and the remaining 10 classes are  used as target (unseen) classes to test the performance of zero-shot classification. The SUN dataset is a fine-grained dataset and contains 717 classes of different scene categories with 20 images per category (14340 images total).  Each image is annotated with 102 attributes that describe the corresponding scene. Following \cite{lampertattribute}, 707 classes are used to learn the dictionaries and the remaining 10 classes are used for testing. The CUB200 dataset is a fine-grained dataset containing 200 classes of different types of birds with 11788 images with 312 attributes and boundary segmentation for each image. The attributes are obtained via human annotation. The dataset is divided into four almost equal folds, where three folds are used to learn the model and the fourth fold is used for testing. For both SUN and CUB200-2011 datasets we used features from VGG19 trained on the ImageNet dataset, which have $4096$ dimensions. We note that our results using ResNet50 and DenseNet \cite{huang2017densely} features will be published in an extended version of this paper. 

 {\bf Tuning parameters:} The optimization regularization parameters $\lambda$, $\rho$, $\gamma$ as well as the number of dictionary atoms $r$ need to be tuned for maximal performance. We used standard $k$-fold cross validation to search for the optimal   parameters for each dataset. After splitting the datasets accordingly into training, validation, and testing sets, we used performance on the validation set for tuning the parameters in a brute-force search. 
we used the common  evaluation metrics in ZSL,  flat hit@K classification accuracy, to measure the performance. This means that a test image is said to be classified correctly  if it is classified among the top $K$ predicted labels. We report hit@1   rate to measure ZSL image classification performance and hit@3 and hit@5 for image retrieval performance. Each experiment is performed ten times and the mean is reported in Tabel \ref{tab:table1}.

{\bf Results:}  Figure \ref{fig:tsne} demonstrates the 2D t-SNE embedding  for predicted attributes and actual class attributes of the AWA dataset.  The actual attributes are depicted by colored circles with black edges. The first column of Figure \ref{fig:tsne} demonstrates the attribute prediction for AAg and AAw formulations. It can be clearly seen that the entropy regularization in AAw formulation improves the clustering quality, decreases data overlap, and reduces the domain shift problem. The nearest neighbor label assignment is shown in the second column, which demonstrates the domain shift and hubness problems with NN label assignment in the attribute space. The third column of Figure \ref{fig:tsne} shows the transductive approach in which a label propagation is performed on the graph of the predicted attributes. Note that the label propagation addresses the domain shift and hubness problem and when used with the AAw formulation provides significantly better zero-shot classification accuracy. 
  
 
Performance comparison results are summarized  in Table \ref{tab:table1}. As pointed out by Xian et al. \cite{xian2017zero} the variety of used image features (e.g. various DNNs and various combinations of these features) as well as the variation of used attributes (e.g. word2vec, human annotation), and different data splits make direct comparison with the ZSL methods in the literature very challenging. In Table \ref{tab:table1} we provide a fair comparison of our JDZSL performance to the recent methods in the literature. All compared methods use the same visual features (i.e. VGG19) and the same attributes (i.e. the continuous or binned) provided in the dataset. 
Table \ref{tab:table1} provides a comprehensive explanation of the shown results.  Note that our method achieves state-of-the-art or close to state-of-the-art performance. 

We report the hit@1 accuracy  on unseen classes in the first nine rows of the table to measure image classification performance. For the sake of transparency and to provide the complete picture to the reader, we included results for the AAg formulation using nearest neighbor, the AAw using nearest neighbor, and AAw using the transductive approach, denoted as transductive attribute aware (TAA) formulation. As it can be seen, while the AAw formulation significantly improves the AAg formulation and adding the transductive approach (i.e. label propagation on predicted attributes) to the AAw formulation further boosts the classification accuracy, as also shown in Figure \ref{fig:tsne}.  In addition, our approach leads to better and comparable performance in all three datasets which include zero-shot scene and object recgonition tasks. More importantly, while the other methods can perform well on a specific dataset,  our algorithm leads to competitive performance on all the three datasets.

\section{Conclusions}
\label{sec:conclusion}
A ZSL formulation is developed that models the relationship between  visual features  and semantic attributes via joint sparse dictionaries. We showed that while a classic joint dictionary learning approach suffers from the domain shift problem, an entropy regularization scheme can help with this phenomenon and provide superior performance. In addition, we demonstrated that a transductive approach towards assigning labels to the predicted attributes can boost the performance considerably and lead to state-of-the-art zero-shot classification.  Finally, we compared our method to the nascent approaches in the literature and demonstrated its competitiveness on benchmark datasets.

\clearpage

\bibliographystyle{aaai}
\bibliography{jointDL}

\clearpage 

\end{document}